\documentclass{article}

\usepackage[final]{neurips_2021}

\usepackage[utf8]{inputenc} % allow utf-8 input
\usepackage[T1]{fontenc}    % use 8-bit T1 fonts
\usepackage{hyperref}       % hyperlinks
\usepackage{url}            % simple URL typesetting
\usepackage{booktabs}       % professional-quality tables
\usepackage{amsfonts}       % blackboard math symbols
\usepackage{nicefrac}       % compact symbols for 1/2, etc.
\usepackage{bbm}

\usepackage{microtype}      % microtypography
\usepackage{xcolor}         % colors
\usepackage{amsmath,amssymb,framed, amsthm}
\usepackage{algorithm}
\usepackage{algorithmic}

\newif\ifcomments
\commentstrue
\ifcomments
    \usepackage{ulem}
    \def\piedit#1{{$\!$\color{magenta} [PI: #1]}}
    
\else 
    \def\piedit#1{}
    
\fi

\ifcomments
    \usepackage{ulem}
    \def\saedit#1{{$\!$\color{blue} [SA: #1]}}
    
\else 
    \def\saedit#1{}
    
\fi

\ifcomments
    \usepackage{ulem}
    \def\rjedit#1{{$\!$\color{brown} [RJ: #1]}}
    
\else 
    \def\rjedit#1{}
    
\fi

\usepackage[mathscr]{eucal}
\usepackage{amsbsy,mathrsfs}
\usepackage{bm}

\include{macrofile2}

\title{Interpretable contrastive word mover's embedding}

\author{%
  Ruijie Jiang\thanks{Optimal Transport and Machine Learning (OTML) workshop, NeurIPS 2021.} \\
  Dept. of Electrical and Computer Engineering\\
  Tufts University\\
  Medford, MA 02155\\
  \texttt{ruijie.jiang@tufts.edu} \\
  
  \And
  Julia Gouvea \\
  Dept. of Education\\
  Tufts University\\
  Medford, MA 02155\\
  \texttt{julia.gouvea@tufts.edu} \\
  
  \And
  Eric Miller\\
  Dept. of Electrical and Computer Engineering\\
  Tufts University\\
  Medford, MA 02155\\
  \texttt{elmiller@ece.tufts.edu} \\
  
  \And
  David Hammer\\
  Dept. of Education\\
  Tufts University\\
  Medford, MA 02155\\
  \texttt{david.hammer@tufts.edu} \\
  \And
  Shuchin Aeron\\
  Dept. of Electrical and Computer Engineering\\
  Tufts University\\
  Medford, MA 02155\\
  \texttt{shuchin@ece.tufts.edu} \\
}

\usepackage{microtype}
\usepackage{amssymb}
\usepackage{amsmath}
\usepackage{booktabs}       % professional-quality tables
\usepackage{amsfonts}       % blackboard math symbols
\usepackage{nicefrac}       % compact symbols for 1/2, etc.
\usepackage{microtype}      % microtypography
\usepackage[utf8]{inputenc}
\usepackage[english]{babel}
\usepackage{multirow}
\usepackage{float}
\usepackage[thinc]{esdiff}

 \usepackage{mathrsfs}
\usepackage{amsmath,amssymb,bm}
\usepackage{graphicx}
\usepackage{algorithm}
\usepackage{algorithmic}
\usepackage{arydshln}
\usepackage{multicol}
\usepackage{verbatim}
\usepackage{tikz}
\usepackage{longtable}

\begin{document}
\maketitle
\begin{abstract}
This paper shows that a popular approach to the supervised embedding of documents for classification, namely, contrastive Word Mover's Embedding, can be significantly enhanced by adding interpretability. This interpretability is achieved by incorporating a clustering promoting mechanism into the contrastive loss. On several public datasets, we show that our method improves significantly upon existing baselines while providing interpretation to the clusters via identifying a set of keywords that are the most representative of a particular class. Our approach was motivated in part by the need to develop Natural Language Processing (NLP) methods for the \textit{novel problem of assessing student work for scientific writing and thinking} - a problem that is central to the area of (educational) Learning Sciences (LS). In this context, we show that our approach leads to a meaningful assessment of the student work related to lab reports from a biology class and can help LS researchers gain insights into student understanding and assess evidence of scientific thought processes. 
\end{abstract}

\section{Introduction}
% \textcolor{red}{Need to check the write-up for anonymity.}

Modern computational methods for Natural Language Processing (NLP) rely on \textit{embeddings} into metric spaces such as the Euclidean, and more recently non-linear spaces such as the Wasserstein space, to achieve state-of-art performance for various  tasks. In these embeddings, the semantic differences and similarities between words and documents, correspond to the distances in the represented space.  For embedding into Euclidean spaces, a large body of work is based on Word2vec~\citep{mikolov2013distributed}, where each word is represented as a vector in the Euclidean space. From these word embeddings one can further compute \textit{document and sentence embeddings} using various models \cite{ramos2003using}, \cite{arora2016simple}, \cite{wang2020sbert}, \cite{le2014distributed}, \cite{kiros2015skip}, \cite{logeswaran2018efficient} for higher level NLP tasks. 

Instead of embedding and comparing documents in the Euclidean space, Word Mover's Distance (WMD) \cite{kusner2015word} was proposed to measure the similarity between documents in the Wasserstein space \cite{COT_book}, representing the documents with (empirical) probability distributions. In \cite{huang2016supervised} WMD is used for supervised learning and more recently in~\citep{yurochkin2019hierarchical}, for multi-scale representation.

To understand how these models work, a lot of effort has been put into aiding interpretability of these embeddings. In \cite{AroraLLMR16} the authors proposed a linear algebraic structure to explain the polysemy of words. Recent works attempted to explain the meaning of each dimension, such as the sparse word embedding \cite{faruqui-etal-2015-sparse, panigrahi-etal-2019-word2sense} and the POLAR Framework \cite{Binny2020}. To make WMD  embeddings interpretable, \cite{Carin2018} proposed an unsupervised \textit{topic model} in the representation space.

In this work our focus is on enabling interpretable \textit{supervised} WMD embeddings of the documents. Below we summarize the main contributions in this direction. 

\textbf{Summary of main contributions} -
A new approach for contrastive representation learning is proposed via enforcing a clustering promoting mechanism using a set of \textit{anchors} that in turn are also learned from the data. This, in contrast to previous approaches \cite{huang2016supervised,kusner2015word}, allows for \emph{interpretability}, i.e. allows one to determine which words are important for a particular class. Furthermore, compared to the $K$ Nearest Neighbour (KNN), our classification using the learned anchors is faster ($\mathcal{O}(n)$ for usual KNN vs $\mathcal{O}(1)$ for our NN using anchors), and our method can be generalized to any other supervised contrastive learning. Results on public data sets as well as a on a novel data set evaluating written scientific work by students show the superiority and utility of our method.

\section{Problem formulation and approach}
% In this paper, we will use lowercase bold to represent vectors, capital bold to represent matrix.

We are given $M$ documents each belonging to one of the $Y$ classes. Each document ${D}^{(m)}$ with label $y^{(m)}$ is represented by two sets, $\{a^m_i\}_{i=1}^{n}$ and $\{b^m_i\}_{i=1}^{n}$, where $n$ is the number of unique words in one document,  $a^m_i$ is the $i$-th word, $b^m_i$ is the number of times $a^m_i$ appears in $D^{(m)}$.  We suppress the dependency of $n$ on $m$ for sake of brevity.

Using  pre-trained  word embeddings from  GLoVe~\cite{pennington2014glove},   $D^{(m)}$ is represented as a tuple $(\bm{X}^{(m)}, \bm{w}^{(m)})$, where $ \bm{X}^{(m)} = [\bm{x}^{(m)}_1, \cdots, \bm{x}^{(m)}_n ] \in \mathbb{R}^{d \times n}$ and $\bm{x}^{(m)}_i \in \mathbb{R}^d$ is the embedding for the word $a^m_i$. The $i$-th entry of $\bm{w}^{(m)} \in \mathbb{R}^{n}$ is $\bm{w}^{(m)}[i]= \frac{b^m_i}{\sum_{i=1}^{n}b^m_i}$, the normalized histogram over the words in the vocabulary.

\textbf{\underline{Problem statement}}: Given labeled data $(\bm{X}^{(m)}, \bm{w}^{(m)}), m = 1,2,\cdots, M$, we seek to learn a representation  $\bm{Z} = f(\bm{X})$ such that a Nearest Neighbor (NN)-type classifier in the representation space accurately classifies the document. 

Using NN in representation space requires a notion of similarity or distances between documents. We use the WMD \cite{kusner2015word}, defined as follows. Given the representations of two documents,  ($\bm{Z}^{(m)}, \bm{w}^{(m)}$), ($\bm{Z}^{(m')}, \bm{w}^{(m')}$) when seen as empirical measures $\mu_m = \sum\limits_{i=1}^{n} \delta_{\bm{z}_i^{(m)}} \bm{w}^{(m)}[i]$, $\nu_{m'} = \sum\limits_{i=1}^{n'} \delta_{\bm{z}_i^{(m')}} \bm{w}^{(m')}[i]$, the WMD between $\mu_{m}$ and $\nu_{m'}$ is defined as \cite{kusner2015word},
\begin{align}
%\label{eq:wmd1}
&\mathsf W(\mu_{m},\nu_{m'}) = \min_{\Gamma} \, \sum_{i,j} d({\bm z_i^{(m)}},{\bm z_j^{(m')}}) \Gamma (i,j) \notag
\end{align}
such that $\Gamma_{i,j} \geq 0$,  $\sum\limits_{i} \Gamma (i,j) = \bm{w}^{(m')}[j]$  and $\sum\limits_{j} \Gamma (i,j) = \bm{w}^{(m)}[i] $. Here $d({\bm z_i^{(m)}},{\bm z_j^{(m')}})$ is referred to as the \textbf{ground cost}. 
% of moving point $\bm z_i^{(m)}$ to $\bm z_j^{(m')}$.

Our \textit{key idea} is to learn a set of \textit{anchors} $\bm{C}^{(y)} \in \mathbb{R}^{d \times p}$ for some $p$ and for each class $y \in  [1: Y]$ in the representation space. Anchors offer two advantages. First, they provide for \textit{direct and simple NN classification}. Second, using anchors we can learn words that have discriminatory power for particular classes, thereby enabling \textit{interpretability}.

% \begin{table*}
% \centering
% {\begin{tabular}{lllllll}
% \hline
% \textbf{DATASET} & description&\#Train & \#Test & Bow dim & avg words & $Y$ \\
% \hline
% \textbf{BBCSPORT} &BBC sports articles labeled by sport & 517& 220&13243&117&5\\
% \textbf{TWITTER} &tweets categorized by sentiment &2176&932&6344&9.9&3\\
% \textbf{RECIPE} &recipe procedures labeled by origin &3059&1311&5703&48.5&15\\
% \textbf{OHSUMED} &medical abstracts (class subsampled) &3999&5153&31789&59.2&10\\
% \textbf{CLASSIC} &academic papers labeled by publisher &4965&2128&24277&38.6&4\\
% \textbf{REUTERS} &news dataset &5485&2189&22425&37.0&8\\
% \textbf{AMAZON} &reviews labeled by product &5600&2400&42063&45.0&4\\
% \textbf{20NEWS} &canonical news article dataset &11293&7528&29671&72&20\\
% \hline
% \end{tabular}}
% \caption{Dataset characteristics.}
% \label{tab:accents}
% \end{table*}

\subsection{Proposed approach}  The representation class for $f$ is defined by$s\bm{A} \in \mathbb{R}^{d \times d}$ and is applied to a document $\bm{X}^{(m)}$ column-wise, 
\begin{align}
\label{eq:model}
    \bm{z}_i^{(m)} = \bm{A} x_i^{(m)}
\end{align} 
where $\circ$ denotes the element-wise product, obtaining a representation $\bm{Z}^{(m)}$. Here, $\bm{A}$ is the transformation matrix.

\begin{figure}[H]
\begin{center}
\includegraphics[width=10cm, height=4cm]{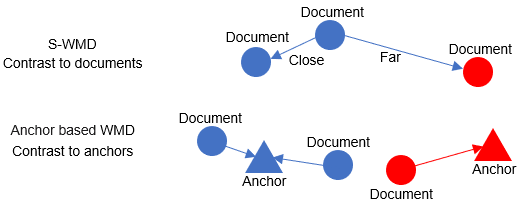}
\end{center}
\vspace{-3mm}
\caption{Schematic illustration of contrastive learning. Different colors means different classes. Top: supervised contrastive learning. Bottom: interpretable contrastive learning with learnable anchors.}
\label{fig:NB}
\vspace{-3mm}
\end{figure}

Given this set-up, our approach is to learn the anchors $\bm{C}^{(y)}, y \in [1:Y]$ via contrastive learning (see Figure \ref{fig:NB}) \cite{chen2020simple,khosla2020supervised} in the representation space. In contrastive learning, one defines triplets $(\mu_m,\mu_{c}^{y_m},\mu_{c}^{y_{m'}})$, where $\mu_m$ is the representation of a document $m$ with label $y_{m}$, $m' \neq m$, and $\mu_{C}^{y_m},\mu_{C}^{y_{m'}}$ are representation of anchors of $\bm{C}^{(y_m)}$ and $\bm{C}^{(y_{m'})}$ respectively. Assuming a uniform measure on the support of the anchor points, we can use the WMD $\mathsf{W}(\mu_m, \mu_{c}^{y_m})$ and $\mathsf{W}(\mu_m, \mu_{c}^{y_{m'}})$ as similarity, i.e. contrastive measure. We will contrast each document with all the anchors. Thus, we will have $Y-1$ triplets $(\mu_m,\mu_{c}^{y_m},\mu_{c}^{y_{m'}})$ for each document. To allow for end-to-end training, we use entropic regularization and use the Sinkhorn algorithm to compute the Wasserstein distance~\cite{cuturi2013sinkhorn}. 

Given $M$ documents, in order to train the model parameters,  $\bm{C}^{(y)}, y \in [1:Y]$ and  $\bm{A}$, we minimize the following \textbf{triplet loss} function~\cite{ hermans2017defense}:
\begin{align}
\label{eq:Loss1}
&\frac{1}{M} \sum\limits_{m=1}^{M} (\sum\limits_{m' \neq m} \max(\mathsf{W}(\mu_m, \mu_{c}^{y_m}) - \mathsf{W}(\mu_m, \mu_{c}^{y_{m'}})) + \beta, 0)) \notag,
\end{align}
where the constant $\beta$ is a margin hyperparameter. We also employ the \textbf{InfoNCE loss} \cite{oord2018representation} to train the model 
%parameters $(\mathbf{A})$:
\begin{equation}
-\frac{1}{M} \sum\limits_{m=1}^{M}  \log(\frac{e^{-\mathsf{W}(\mu_m, \mu_{c}^{y_m})/\tau}}{e^{- \mathsf{W}(\mu_m, \mu_{c}^{y_m})/\tau}+\sum\limits_{m \neq m'} e^{- \mathsf{W}(\mu_m, \mu_{c}^{y_{m'}})/\tau}}), \notag
\end{equation}
where $\tau$ denotes a temperature parameter.

\textbf{NN classification}: Once the model is trained, in order to do NN classification on a test document, we first embed it in the representation space using the learned parameters, $\bm{A}$ via equation \eqref{eq:model}, compute the WMD distances between the representation and the anchors $\bm{C}^{(y)}, y \in [1:Y]$. The class represented by the anchor with the minimum distance is declared as the label.

\textbf{Interpretability}: To show how one can use the anchors to discover important discriminative words, we refer the reader to section \ref{sec:interpretation} where we illustrate it with a concrete example.
\section{Evaluation} 
For all datasets, the ground cost in WMD is set to squared Euclidean and we use the Sinkhorn algorithm for computing it \cite{COT_book}. The various hyper-parameters are set using cross-validation.  For the triplet loss, the margin is set to $\beta = 10$, and for the InfoNCE loss the temperature parameter is set to $\tau=30$.  We used the Adam~\cite{kingma2014adam} optimizer with learning rate of 0.1. We employed the pre-trained GLoVe~\cite{pennington2014glove} word vectors with dimension $d=300$ as the representation of words. To avoid overfitting, we employed $\ell_2$ regularization with parameter $0.001$.

\begin{table*}[h!]
\centering
\begin{tabular}{lllllll}
\hline
\textbf{DATASET} &\#Train & \#Test & Bow dim & avg words & $Y$ \\
\hline
\textbf{BBCSPORT} & 517& 220&13243&117&5\\
\textbf{TWITTER} &2176&932&6344&9.9&3\\
\textbf{RECIPE} &3059&1311&5703&48.5&15\\
\textbf{OHSUMED}  &3999&5153&31789&59.2&10\\
\textbf{CLASSIC}  &4965&2128&24277&38.6&4\\
\textbf{REUTERS} &5485&2189&22425&37.0&8\\
\textbf{AMAZON}  &5600&2400&42063&45.0&4\\
\textbf{20NEWS}  &11293&7528&29671&72&20\\
\hline
\end{tabular}
\caption{Public dataset characteristics.}
\label{tab:accents}
\end{table*}

\textbf{Public Datasets}:\footnote{\textbf{Note}: The public dataset can be found https://github.com/gaohuang/S-WMD}
Information about the public datasets is shown in Table 1.  In Table 2 we show the results from WMD \cite{kusner2015word} , supervised-WMD (S-WMD) \cite{huang2016supervised} and our method. We can see our method successfully outperformed WMD and S-WMD in seven out of the eight datasets. 
% We notice that the representation of documents with the same labels are not always close to each other so that the performance of a typical KNN classifier is poor. On the other hand using the anchors for NN classification yields stronger performance.

% \begin{figure}
% \begin{center}
% \includegraphics[width=8cm, height=5cm]{T-SNE.png}
% \end{center}
% \caption{t-SNE plots on BBC sports}
% \label{fig:NB}
% \end{figure}
% \begin{figure}
% \begin{center}
% \includegraphics[width=7cm, height=6cm]{TSNE_class.png}
% \end{center}
% \caption{t-SNE plots of each classes representation on BBC sports, each point correspond to a support point.}
% \label{fig:NB}
% \end{figure}
\begin{table*} 
\scriptsize
\centering
\begin{tabular}{lllllllll}
\hline
\textbf{DATASET} & \textbf{BBCSPORT} & \textbf{TWITTER} & \textbf{RECIPE} & \textbf{OHSUMED} & \textbf{CLASSIC} & \textbf{REUTERS} & \textbf{AMAZON} & \textbf{20NEWS}\\
\hline
WMD & $4.6\pm0.7$ & $28.7 \pm 0.6$ & $42.6 \pm 0.3$ & $44.5$ & $\mathbf{2.8 \pm 0.1}$ & $3.5$ &$7.4 \pm 0.3$ & $28.3$\\
S-WMD & $2.1\pm0.5$ & $27.5 \pm 0.5$ & $39.2 \pm 0.3$ & $34.3$ & $3.2 \pm 0.2$ & $3.2$ &$5.8 \pm 0.1$ & $26.8$\\
\hdashline
\underline{Ours} - \textbf{triplet loss} & $\mathbf{1.7\pm0.4}$ & $\mathbf{24.8 \pm 0.8}$ &$39.2 \pm 0.4$ & $\mathbf{33.0}$ &$3.0 \pm 0.3$ &$\mathbf{2.8}$ & $5.6 \pm 0.5$ & $\mathbf{26.7}$\\
\underline{Ours} - \textbf{InfoNCE loss} & ${2.4\pm0.6}$ & ${25.4 \pm 1}$ &$\mathbf{38.6 \pm 0.4}$ & ${33.5}$ &$3.4 \pm 0.4$ &${2.9}$ &$\mathbf{5.5 \pm 0.4}$ & $26.8$\\
\hline
\end{tabular}
\caption{\label{a-guide}
Classification error rate for our methods and other baselines.
}
\end{table*}

\subsection{Interpretation} 
\label{sec:interpretation}
\begin{figure*}
\begin{center}
\includegraphics[width=16cm, height=7cm]{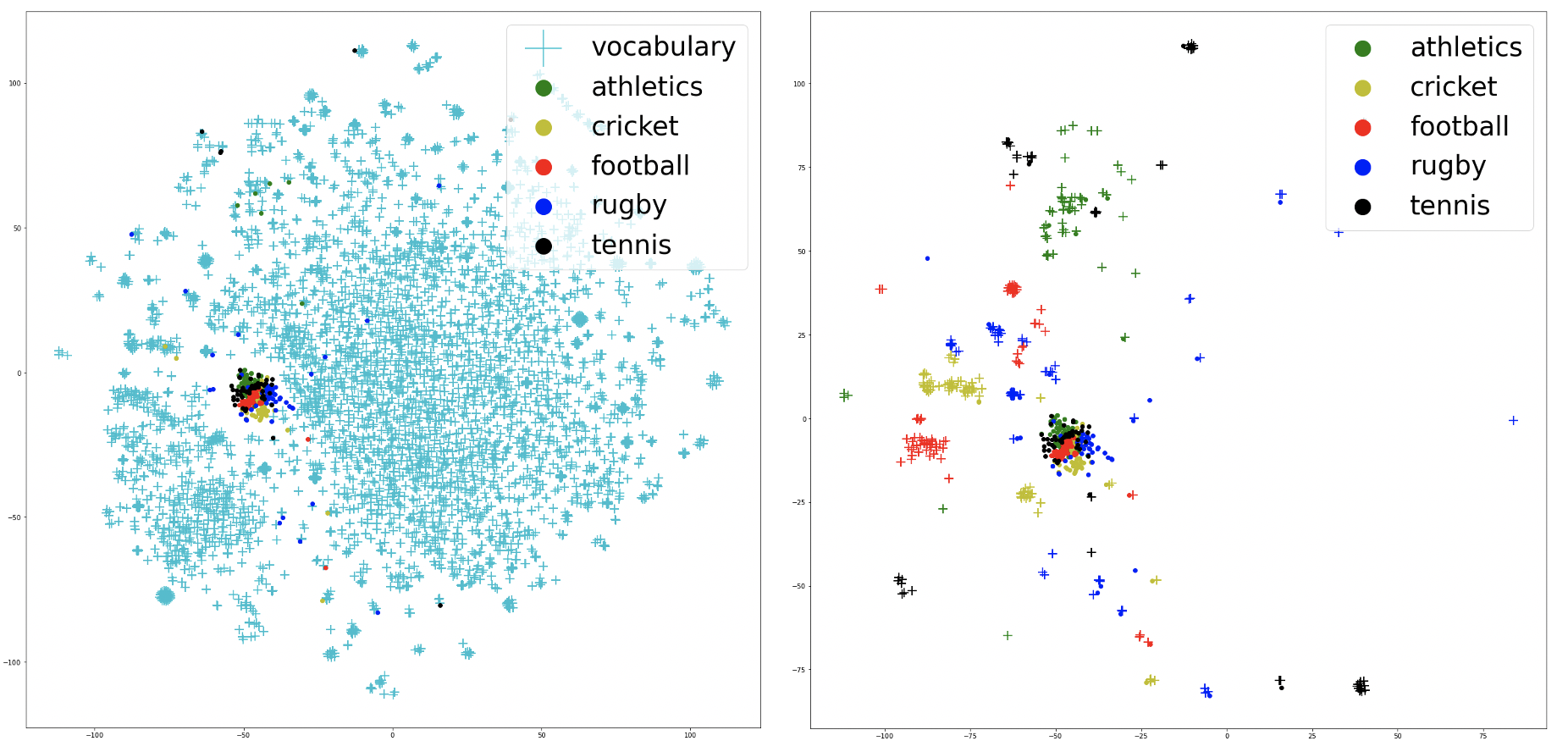}
\end{center}
\vspace{-4mm}
\caption{t-SNE visualization of vocabulary and support points for class representation. Left panel shows all vocabulary and class representation, right panel shows most important words and class representation.}
\vspace{-5mm}
\label{fig:tSNE}
\end{figure*}

Figure \ref{fig:tSNE} shows how our model leads to interpretability. Under the contrastive loss, the difference between WMD $\mathsf W(\mu_m, \mu_c^{y_m})$ and $\mathsf W(\mu_m, \mu_c^{y_{m'}})$ will be maximized. This forces the \textit{important words} for a given class to be close to the anchor of its corresponding class in the representation space and further from the anchors of the other classes. Also, the \textit{common words} for all classes will have a relatively similar distance to any of the anchors as they play no role in discrimination. Concretely, for the learned word representation $\bm z_t$ of word $a_t$ (using $\bm{A}$) and the anchor $\bm{C}^{(y)}$, we  define the distance $\mathsf D(\bm z_t, \bm{C}^{(y)}) = \min \{d(\bm z_t,\bm{q}^{(y)}_i), i = 1, \cdots, p\}$, where $p$ is the number of support points and $\{\bm{q}_i^{(y)}\}_{i=1}^{p}$ are columns (support points) of the anchor $\bm{C}^{(y)}$. Then we define the importance value of $\bm{z}_t$ for class $y$ as  $I(\bm z_t,y) = \sum_{k\neq y}\mathsf D(\bm z_t, \bm{C}^{(k)}) - (Y-1)\times \mathsf D(\bm z_t, \bm{C}^{(y)})$. Larger $I(\bm z_t,y)$ means that word $a_t$ is important for class $y$. The basic idea behind the interpretation is that the learned anchor for each class can be understood  as an “abstract document” for this class in the representation space. We believe that the overlap between different anchors is something that is in common for different classes. For a given class, the non-overlapping parts (with other classes) can be viewed as the important features in the representation space for this class, and we show that the words close to the non-overlapping part in the representation space are indeed the important words for a given class.

% If $I(\bm z_t,y)$ is large, it requires $\mathsf D(\bm z_t, C^{(y)})$ to be small and $\sum_{k\neq y}\mathsf D(\bm z_t, \bm{C}^{(k)})$ to be large, which means word $a_t$ is more related to class $y$. Thus, the word $ a_t$ with large $I(\bm z_t,y)$ can be viewed as the important word.\\

In Figure \ref{fig:tSNE}, we show the t-SNE visualization of learned anchors and vocabulary. We see the anchors corresponding to different classes have some overlap. But, more importantly, as shown in the right panel of Figure \ref{fig:tSNE}, the important words for each class generated from our method are \textit{not} overlapping and are relatively far from each other. 

% We see the anchors corresponding to different classes have some overlap. This makes sense since for each sentence, there should be a certain ratio of unimportant words which do not provide information for classification.

We show the top-30 words for each class from BBCsports dataset in Figure \ref{fig:topBBC}. We checked the frequency of these words and we believe these words are important. For example, the Top-30 words we generated for "Cricket" were totally shown $66$ times in the dataset, and $64$ of them belong to the class ``Cricket''.

\begin{figure}
\begin{center}
\includegraphics[width=12cm, height=6.5cm]{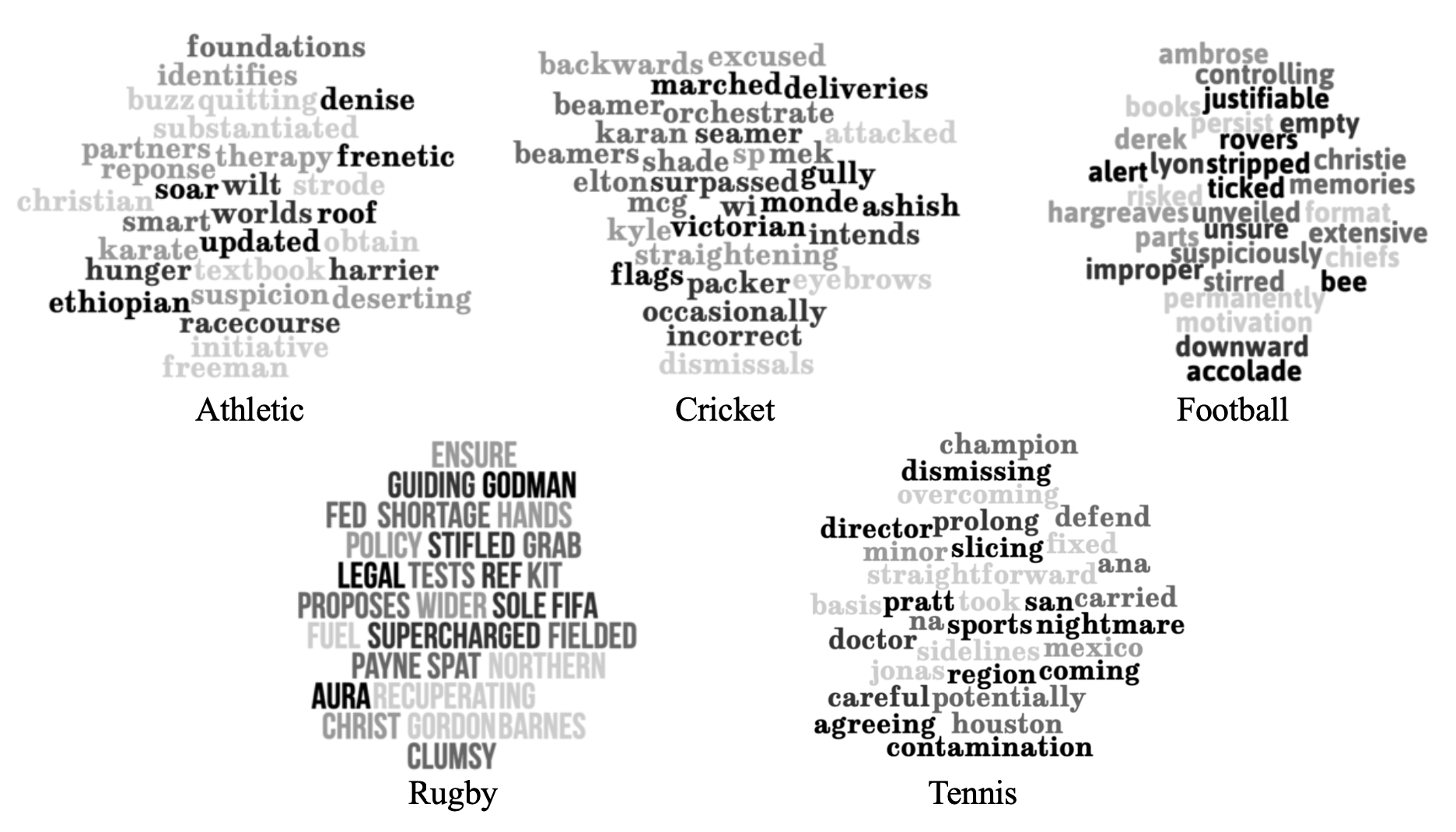}
\end{center}
\vspace{-3mm}
\caption{The Top-30 words of each class on the BBCSports dataset.}
\vspace{-3mm}
\label{fig:topBBC}
\end{figure}

\textbf{Assessing written student work}: 
This work concerns analyses by an instructor to track shifts in the quality of students’ writing due to curricular innovations.  An overview of the dataset is shown in Table \ref{tab:lab_data}. 

Human coders separated lab reports into higher and lower scores using an adapted version of the SOLO taxonomy (Biggs \& Collis, 1982), which uses three features of reports to determine quality: claim complexity, scope of evidence, and consistency and closure (Authors, in prep).  

For this dataset, since we have rather limited data to work with, we first combine the scores (1,2) as a low score, and scores (3,4) as a high score, yielding a binary classification task.
\begin{table*} [h!]
\scriptsize
\centering
\begin{tabular}{lllllllll}
\hline
Essays & Avg length & Max length & Min length & Scores\\
\hline
146 & 512 & 1449 & 95 & 1-4\\ 
\hline
\end{tabular}
\caption{Overview of the lab report dataset.\label{tab:lab_data}}
\end{table*}

\begin{figure}
\begin{center}
\includegraphics[width=10cm, height=5cm]{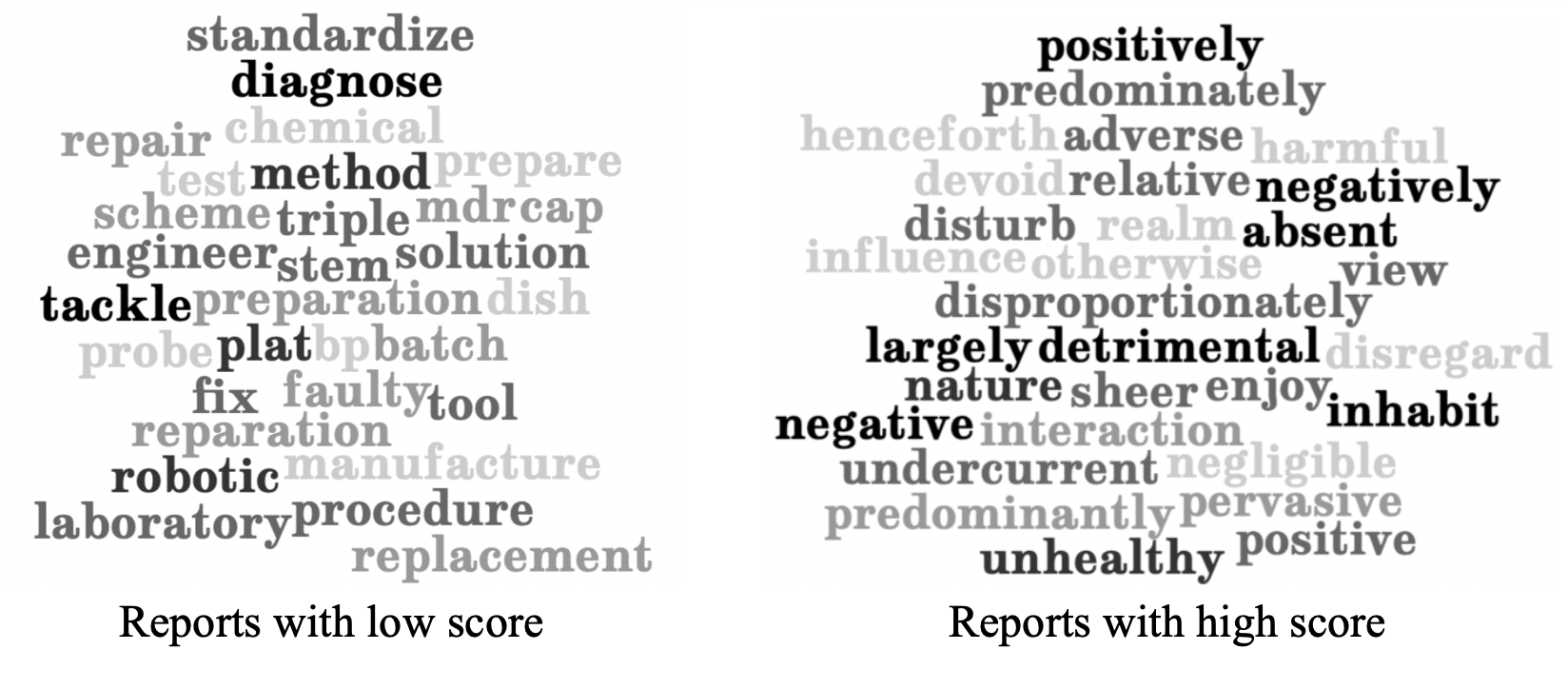}
\end{center}
\vspace{-3mm}
\caption{The Top-30 words of each class for the lab reports dataset generated by our method.}
\vspace{-3mm}
\label{fig:topBio}
\end{figure}

\begin{figure}[htbp!]
\begin{center}
\includegraphics[width=8cm, height=4.2cm]{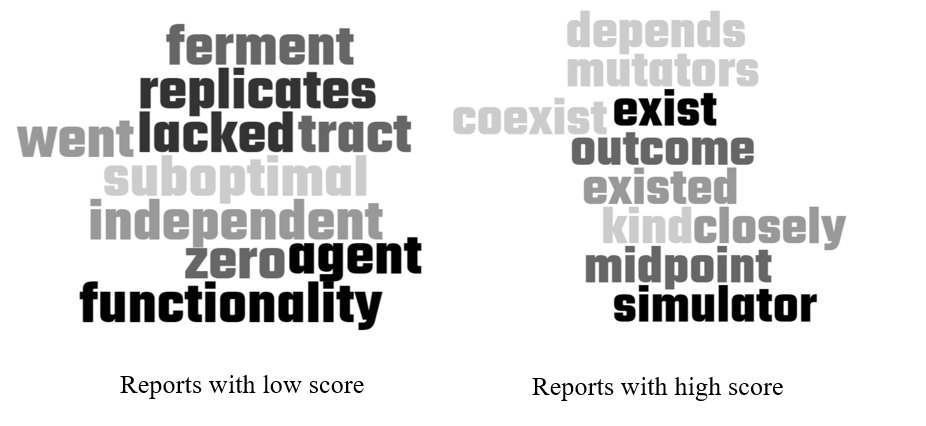}
\end{center}
\vspace{-3mm}
\caption{The Top-10 words of each class for the lab reports dataset generated by TF-IDF.} \label{TF-IDF}
\vspace{-3mm}
\label{fig:topBBC}
\end{figure}

\underline{\textit{Performance and interpretation}}: In Table \ref{a-guide2} shows the accuracy of the three methods.
Our method, which improves on WMD, is not quite as accurate as S-WMD neither of which provide for \textit{inerpretatability}, a defining benefit of the work in this paper. The performance gap is due to the fact that our method requires learning more parameters, namely the anchors, compared to S-WMD. Since the training data is rather limited for the new dataset, our error rate is higher. This issue is not present in the public dataset. In Figure \ref{fig:topBio}, we plot the top 30 words for reports with low and high scores. To make comparison, we also list top words generated by TF-IDF for lab reports with low score and high score separately in Figure \ref{TF-IDF}.

\begin{table*} [h!]
\scriptsize
\centering
\begin{tabular}{lllllllll}
\hline
\textbf{Method} &\textbf{WMD} &\textbf{S-WMD} &\textbf{Ours (triplet loss)}\\
\hline
\textbf{Error rate} &$21.4 \pm 0.6$  &$16.6 \pm 0.3$& $20  \pm 0.4$\\
\hline
\end{tabular}
\caption{\label{a-guide}
Error rate for lab report data\label{a-guide2}}
\end{table*}

\underline{\textit{Discussion of lab report results}}: 
The discriminatory words identified by our approach, suggest a good fit with the qualitative differences, namely, claim complexity, scope of evidence, and consistency and closure, used by human coders to make classifications. The words also suggest themes not directly coded for. 

For example, differences in adjectives reflect differences in claim structure. The importance of adjectives such as positive, negative and relative, reflect the more complex claim structure in high scoring reports. While low scoring reports stated simple claims, high scoring reports compared the relative influence of competing effects (i.e. positive and negative mutations).

Another hallmark of high scoring report is qualified or conditional claims that indicate context-specificity or uncertainty. The importance of adverbs such as predominantly, largely, and disproportionately, in high-scoring reports, reflects uncertainty, expressed as of probabilistic claims, that were common in these reports. While these properties were not observed from the top words generated by TF-IDF.

The predominance of nouns and verbs that describe laboratory procedures (e.g. method, procedure, standardize) in low-scoring reports is an interesting difference not directly coded for by human coders. It is nevertheless consistent with the shift in the laboratory curriculum from an emphasis on reporting on procedures to interpreting and arguing about findings that underlies the shift from low to high scores.

Overall these findings suggest that \textit{our method captures meaningful qualitative differences} originally identified by qualitative researchers. 

\section{Codes and Implementation}
 Our code can be found at https://github.com/rjiang03/Interpretable-contrastive-word-mover-s-embedding.

% \section{Conclusions and future work}
% \textcolor{red}{Not sure if we need this for review purposes.}

% \section{Appendices}

% Use \verb|\appendix| before any appendix section to switch the section numbering over to letters. See Appendix~\ref{sec:appendix} for an example.

\section*{Acknowledgements}
This research is supported by NSF RAISE 1931978. Shuchin Aeron is aso supported in part by NSF CCF:1553075, NSF ERC planning 1937057, and AFOSR FA9550-18-1-0465. 
\newpage

% \bibliographystyle{abbrev}
\bibliographystyle{ACM-Reference-Format}
\bibliography{emnlp2021}
% \bibliographystyle{acl_natbib}

% \appendix

% \section{Example Appendix}
% \label{sec:appendix}

% This is an appendix.

\end{document}